\documentclass[11pt]{article}
\usepackage{amsmath, amssymb}
\usepackage{graphicx}
\usepackage{hyperref}
\usepackage{natbib}

\usepackage{float}

\title{Decomposer Networks:\\ Deep Component Analysis and Synthesis}
\author{Mohsen Joneidi\\ mohsen.joneidi@outlook.com}
\date{}

\begin{document}
\maketitle

\begin{abstract}
We propose the \textit{Decomposer Networks} (DecompNet), a semantic autoencoder that 
factorizes an input into multiple interpretable components. Unlike classical 
autoencoders that compress an input into a single latent representation, the 
Decomposer Network maintains $N$ parallel branches, each assigned a residual 
input defined as the original signal minus the reconstructions of all other 
branches. By unrolling a Gauss--Seidel style block-coordinate descent into a 
differentiable network, DecompNet enforce explicit competition among 
components, yielding parsimonious, semantically meaningful representations. 
We situate our model relative to linear decomposition methods (PCA, NMF), 
deep unrolled optimization, and object-centric architectures (MONet, IODINE, 
Slot Attention), and highlight its novelty as the first semantic autoencoder 
to implement an \textit{all-but-one residual update rule}.
\end{abstract}

\section{Introduction}
Human creativity often begins with decomposition: breaking complex experiences into their essential parts. A skilled chef separates flavors, a painter distinguishes tones and textures, a musician isolates harmonies, and a mathematician dissects structures into simpler forms. This ability to analyze the whole through its components is at the heart of deep understanding. In mathematics, the singular value decomposition—introduced almost 150 years ago—embodied this principle, providing a powerful way to separate a matrix into fundamental elements with elegant and useful properties. Today, as we enter the era of artificial intelligence, the challenge is to equip machines with a comparable capacity for structured, component-wise reasoning. To this end, we introduce Decomposer Networks, a neural architecture designed to extend the spirit of SVD into the nonlinear and semantic domain of AI.

Decomposition of data into semantic components is a longstanding goal in 
signal processing and representation learning. Classical methods such as PCA 
and NMF provide additive factorization but are restricted to linear settings. 
Autoencoders and variational autoencoders capture nonlinear structure, yet 
entangle semantics within a single latent vector. Object-centric models 
introduce multi-slot representations, but rely on masking and attention 
rather than residual explain-away.

We propose DecompNet, a semantic autoencoder that 
assigns each branch a residual view of the input, enforcing specialization 
and interpretability. This architecture bridges the gap between explain-away 
principles from sparse coding and modern deep neural factorization. Related work are listed below:

\paragraph{Linear and Shallow Decomposition.}
Classical approaches such as PCA, ICA, and NMF 
\citep{jolliffe2002principal, hyvarinen2000independent, lee1999learning} 
provide additive decompositions but remain linear.

\paragraph{Deep Unrolled Factorization.}
Works such as LISTA \citep{gregor2010learning}, ADMM-Net \citep{yang2016admm}, 
and deep NMF \citep{trigeorgis2016deep} unroll optimization into neural updates. 
They lack the residual competition mechanism we propose.

\paragraph{Object-Centric Scene Decomposition.}
Models such as MONet \citep{burgess2019monet}, IODINE \citep{greff2019iodine}, 
and Slot Attention \citep{locatello2020object} decompose inputs into slots 
using masking and attention. Our method instead employs residual subtraction 
to enforce explain-away dynamics.

\paragraph{Residual Factorization in Networks.}
Factorized residual units \citep{chen2017sharing} improve efficiency, but 
focus on parameter sharing rather than semantic decomposition.

\section{Relation to Classic Decompositions}
\label{sec:relation_to_classic}
To highlight the connection between Decomposer Networks and classical 
linear factorization, we consider a simplified setting in which each branch 
is a purely linear operator, i.e. $F_i(r) = W_i r$ and $S_i(y_i) = V_i y_i$, 
with $W_i \in \mathbb{R}^{k \times d}$ and $V_i \in \mathbb{R}^{d \times k}$.
The overall reconstruction after one sweep is therefore
\begin{equation}
    \hat{x} = \sum_{i=1}^N V_i W_i r_i,
\end{equation}
where the residual $r_i$ is defined as $x$ minus the reconstructions of the 
other components.

\paragraph{Rank-one initialization.}
Assume each branch is initialized as a rank-one linear operator:
\[
    V_i W_i = u_i v_i^\top,
\]
where $u_i, v_i \in \mathbb{R}^d$ are drawn at random with unit norm. 
Thus, each branch initially captures a one-dimensional projection of the input.

\paragraph{Iteration dynamics.}
During Gauss--Seidel sweeps, branch $i$ is updated on the residual
\[
    r_i^{(t)} = x - \sum_{j \neq i} u_j v_j^\top x,
\]
which ensures that $r_i^{(t)}$ lies in the orthogonal complement of the 
subspace spanned by $\{u_j v_j^\top\}_{j \neq i}$ up to reconstruction error. 
Applying $v_i^\top$ extracts the dominant direction in that residual, and 
updating $u_i$ aligns it with $r_i^{(t)}$.

\paragraph{Connection to SVD.}
This procedure is mathematically equivalent to \textit{deflation methods} 
for singular value decomposition. Classical deflation iteratively subtracts 
rank-one approximations from a matrix or signal until convergence, with 
each step converging to the next singular component 
\citep{golub2013matrix}. In our setting, each Gauss--Seidel update
performs an analogous step: the first branch converges to the dominant 
singular component $(u_1, v_1)$, the second to the next $(u_2, v_2)$, and 
so forth. After sufficient iterations, the collection of branches recovers 
the SVD of the input up to scaling and ordering.

\paragraph{Implication.}
Therefore, Decomposer Networks can be viewed as a \textit{nonlinear extension} 
of SVD. In the linear case with rank-one subnetworks, they reduce to 
classical singular value decomposition via iterative deflation. In the 
general nonlinear case, they retain the explain-away residual dynamics but 
extend beyond linear manifolds, enabling decomposition into semantic 
components that need not be orthogonal or linear.

\begin{figure}
    \centering
    \includegraphics[width=0.95\linewidth]{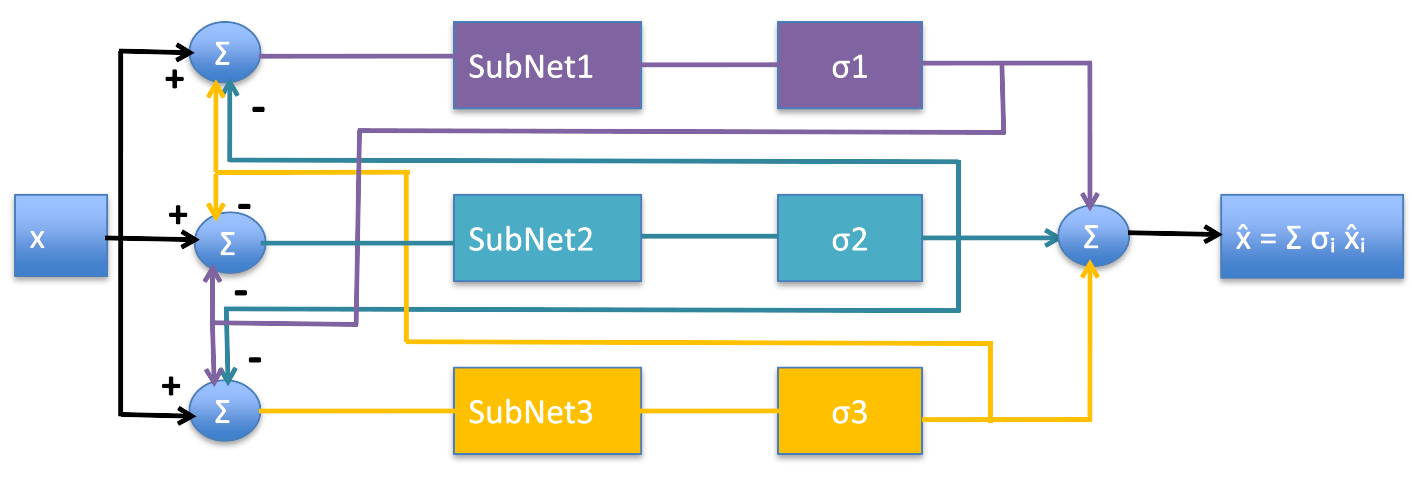}
    \label{fig:block_diag}
\caption{\small{Decomposer Networks (3 components). Each residual summer adds \(x\) and subtracts the \emph{other} branches’ scaled reconstructions (\(-\sigma_j\hat{x}_j\)). Each color shows one component and colored arrows show components data; gains \(\sigma_i\) feed both the final sum and the residual feedback. Each SubNet can be as simple as a rank-1 multiplication or as deep as a multi layer auto-encoder.}}
\end{figure}

\section{Model and Cost Function}
Given input $x \in \mathbb{R}^d$, the network learns $N$ components 
$\{y_i\}_{i=1}^N$ via branch-specific encoders $F_i$ and decoders $S_i$:

\begin{equation}
    \hat{x}_i = S_i(y_i), \qquad \hat{x} = \sum_{i=1}^N \hat{x}_i.
\end{equation}

Each branch $i$ receives as input a residual defined by the reconstructions 
of all other branches:
\begin{equation}
    r_i^{(t)} = x - \sum_{j \neq i} \hat{x}_j^{(t)}.
\end{equation}

The branch update is then:
\begin{equation}
    y_i^{(t)} = F_i(r_i^{(t)}), \qquad 
    \hat{x}_i^{(t)} = S_i(y_i^{(t)}).
\end{equation}

Iterating over $i=1,\ldots,N$ for $K$ sweeps yields a Gauss--Seidel style 
residual refinement. Training minimizes a composite loss:
\[
\mathcal{L} = \|x - \hat{x}\|^2 
+ \lambda_s \sum_i \|y_i\|_1 
+ \lambda_\perp \sum_{i\neq j}\langle \hat{x}_i, \hat{x}_j\rangle^2,
\]
with optional semantic heads to align each component to supervised labels.

Decomposer Networks are the first semantic autoencoders to implement an 
explicit \textit{all-but-one residual update rule}. Each branch is forced to 
model what the others cannot, producing semantic disentanglement by design. 
Compared to deep unrolled methods, our updates are residual-conditioned and 
sequential; compared to object-centric models, our decomposition arises from 
residual explain-away rather than attention masks.

\section{Optimization and Learning}
\label{sec:optimization}

\paragraph{Setup.}
We are given a dataset $\mathcal{D}=\{x^{(n)}\}_{n=1}^B$ (mini-batch size $B$).
The Decomposer Network contains $N$ \emph{autoencoders} (AEs).
AE $i$ has an encoder $E_i(\cdot;\theta_i)$ and decoder $D_i(\cdot;\phi_i)$ producing a component reconstruction
\[
\hat{x}_i \;=\; D_i\!\big(E_i(r_i)\big),
\]
where $r_i$ is the \emph{residual input} to AE $i$ defined by
\begin{equation}
\label{eq:residual}
r_i \;=\; x \;-\! \sum_{j\neq i} \sigma_j\, \hat{x}_j.
\end{equation}
The final reconstruction is the scaled sum
\begin{equation}
\label{eq:final-sum}
\hat{x} \;=\; \sum_{i=1}^{N} \sigma_i\, \hat{x}_i,
\end{equation}
where $\boldsymbol{\sigma}=[\sigma_1,\dots,\sigma_N]^\top$ are \emph{per-sample} nonnegative scalars (analogous to singular values in SVD). We optionally perform $K$ Gauss--Seidel sweeps over $i$ to refine $\{\hat{x}_i\}$ (weights tied across sweeps unless otherwise noted).

\subsection{Objective}
For a mini-batch, we minimize
$$
\mathcal{L} \;=\; 
\frac{1}{B}\sum_{n=1}^{B} 
\underbrace{\big\|x^{(n)} - \sum_{i}\sigma^{(n)}_i\,\hat{x}^{(n)}_i\big\|_2^2}_{\text{reconstruction}}$$
\begin{equation}
\label{eq:loss}
\;+\; \lambda_s \sum_{i}\|\mathbf{z}_i\|_{1}
\;+\; \lambda_{\perp}\sum_{i\neq j}\!\!\big\langle \hat{x}_i,\hat{x}_j\big\rangle^2
\;+\; \sum_i \lambda_{\text{sem},i}\,\mathcal{L}_{\text{sem},i},
\end{equation}
where $\mathbf{z}_i=E_i(r_i)$ are AE codes (sparsity promotes parsimony), the orthogonality/independence penalty $\lambda_{\perp}$ reduces component overlap, and $\mathcal{L}_{\text{sem},i}$ are optional semantic heads if supervision is available. All inner products, norms, and losses are computed per-sample then averaged over the batch.

\subsection{Per-sample scaling coefficients \texorpdfstring{$\boldsymbol{\sigma}$}{sigma}}
\label{sec:sigma}
For a fixed set of component reconstructions $\{\hat{x}_i\}_{i=1}^N$ (produced by the current AEs), the optimal scaling $\boldsymbol{\sigma}$ for \emph{each} sample $x$ solves a small nonnegative least-squares (NNLS):
\begin{equation}
\label{eq:sigma-ls}
\min_{\boldsymbol{\sigma}\ge 0}\;\;\big\|x - \mathbf{H}\,\boldsymbol{\sigma}\big\|_2^2,
\qquad \text{with}\quad
\mathbf{H}=[\hat{x}_1\,\,\hat{x}_2\,\,\cdots\,\,\hat{x}_N]\in\mathbb{R}^{d\times N}.
\end{equation}
When nonnegativity is not enforced, the closed-form is
\begin{equation}
\label{eq:sigma-closed}
\boldsymbol{\sigma}^\star \;=\; (\mathbf{H}^\top \mathbf{H} + \varepsilon \mathbf{I})^{-1}\mathbf{H}^\top x,
\end{equation}
with a tiny Tikhonov $\varepsilon>0$ for stability. With NNLS, a fast projected gradient or active-set solver suffices because $N$ is small. Importantly, we compute \eqref{eq:sigma-closed} (or NNLS) \emph{independently for each sample} in the batch.

\subsection{Alternating training (residual coordinate descent)}
We alternate between updating the per-sample scalars \(\boldsymbol{\sigma}\) and updating AE weights \(\{\theta_i,\phi_i\}\). Each outer iteration uses one mini-batch.

\paragraph{Step A: Update \(\boldsymbol{\sigma}\) (for each sample).}
Hold AE weights fixed. For each $x^{(n)}$, compute current components $\hat{x}^{(n)}_i$ by \eqref{eq:residual}–\eqref{eq:final-sum} (one or more Gauss--Seidel sweeps), form $\mathbf{H}^{(n)}$, then solve \eqref{eq:sigma-ls} (NNLS) or \eqref{eq:sigma-closed} to obtain $\boldsymbol{\sigma}^{(n)}$.

\paragraph{Step B: Update AE weights (one or more sweeps).}
Hold $\{\boldsymbol{\sigma}^{(n)}\}$ fixed. For each sweep $t=1,\dots,K$ and branch $i=1,\dots,N$:
\begin{align}
r_i^{(n,t)} &= x^{(n)} - \!\!\sum_{j\neq i}\!\! \sigma^{(n)}_j \,\hat{x}_j^{(n,t)} \quad \text{(Gauss--Seidel uses latest $\hat{x}_j$)}, \\
\mathbf{z}_i^{(n,t)} &= E_i\!\big(r_i^{(n,t)};\theta_i\big),\qquad 
\hat{x}_i^{(n,t)} = D_i\!\big(\mathbf{z}_i^{(n,t)};\phi_i\big).
\end{align}
Accumulate the batch loss \eqref{eq:loss} with $\hat{x}=\sum_{i}\sigma^{(n)}_i \hat{x}_i^{(n,t)}$ and update $\{\theta_i,\phi_i\}$ by backpropagation (any first-order optimizer). Optionally use \emph{relaxation} (damping) to improve stability:
\[
\hat{x}_i^{(n,t)} \leftarrow (1-\alpha)\,\hat{x}_i^{(n,t-1)} + \alpha\,\hat{x}_i^{(n,t)},\qquad \alpha\in(0,1].
\]

\paragraph{Jacobi vs.\ Gauss--Seidel.}
Jacobi updates compute all $r_i$ from the previous sweep (parallelizable on GPUs); Gauss--Seidel consumes freshest neighbors (often faster empirical convergence). Both are differentiable end-to-end.

\subsection{Algorithm}
\begin{center}
\begin{minipage}{0.93\linewidth}
\begin{flushleft}
\textbf{Algorithm 1:} Alternating training for Decomposer AEs with per-sample $\boldsymbol{\sigma}$\\
\textbf{Input:} batch $\{x^{(n)}\}_{n=1}^B$, AEs $\{E_i,D_i\}$ with weights $\{\theta_i,\phi_i\}$, sweeps $K$, damping $\alpha$.\\
\textbf{Repeat until convergence:}
\begin{enumerate}
\item \textbf{(Forward components)} For each $n$: initialize $\hat{x}_i^{(n,0)}\!=\!0$.\\
\hspace*{1.3em}For $t=1..K$, for $i=1..N$: form $r_i^{(n,t)}$ by \eqref{eq:residual}, compute $\hat{x}_i^{(n,t)}=D_i(E_i(r_i^{(n,t)}))$, optionally relax with $\alpha$.
\item \textbf{(Per-sample scales)} For each $n$: form $\mathbf{H}^{(n)}=[\hat{x}_1^{(n,K)},\dots,\hat{x}_N^{(n,K)}]$ and solve \eqref{eq:sigma-ls} (or \eqref{eq:sigma-closed}) for $\boldsymbol{\sigma}^{(n)}$.
\item \textbf{(Backprop AEs)} With $\{\boldsymbol{\sigma}^{(n)}\}$ fixed, recompute residual sweeps and minimize \eqref{eq:loss} by SGD/Adam w.r.t.\ $\{\theta_i,\phi_i\}$.
\end{enumerate}
\textbf{Output:} trained $\{\theta_i,\phi_i\}$ and, at inference, per-sample $\boldsymbol{\sigma}$ via Step~2.
\end{flushleft}
\end{minipage}
\end{center}

\subsection{Gradients and practical notes}
\noindent\textbf{Backprop through residuals.}
In Step~B, $\boldsymbol{\sigma}$ is held fixed; gradients flow through the residual construction \eqref{eq:residual} and through each AE. Because $r_i$ depends on $\hat{x}_j$, a branch update indirectly influences others, which is precisely the desired competitive coupling.

\noindent\textbf{Nonnegativity and normalization.}
Enforce $\sigma_i\ge 0$ either by NNLS, by a softplus parameterization $\sigma_i=\mathrm{softplus}(\tau_i)$, or by projecting negative values to zero after \eqref{eq:sigma-closed}. To avoid trivial rescalings, apply weight normalization to decoders or constrain $\|\hat{x}_i\|_2$ (e.g., divide by its norm inside $\mathbf{H}$ and absorb scale into $\sigma_i$).

\noindent\textbf{Stability.}
Use small $K$ initially (e.g., $K=1$), then increase to $3$–$5$. Damping $\alpha\in[0.3,0.7]$ reduces ``ping–pong'' between branches. Orthogonality/independence penalties ($\lambda_{\perp}$) curb duplicate explanations.

\noindent\textbf{Permutation symmetry.}
To prevent slot swapping, bias branches slightly differently (e.g., distinct receptive fields or weak semantic heads), or add mild diversity priors.

\noindent\textbf{Inference.}
At test time, compute components via $K$ sweeps and estimate $\boldsymbol{\sigma}$ per sample by \eqref{eq:sigma-ls}; report both $\{\hat{x}_i\}$ and $\{\sigma_i\}$.

Potential use cases of DecompNet include:
\begin{itemize}
    \item Time-series decomposition (trend, oscillatory modes, noise)
    \item Radar/communications (clutter vs. target vs. multipath separation)
    \item Images (structure vs. texture vs. illumination)
    \item Biomedical signals (e.g., ECG/EEG component separation)
\end{itemize}

\section{Experimental Results}

\subsection{Dataset}
All experiments were conducted on the \textbf{AT\&T Faces Dataset} (formerly known as the ORL database)~\cite{samaria1994att}. The dataset contains 400 grayscale images of 40 subjects, each with variations in facial expression, pose, and illumination. Each image has an original resolution of $112\times92$ pixels, which was optionally downsampled to $56\times46$ for computational efficiency. All images were standardized to zero mean and unit variance per feature prior to training.

\subsection{Experiment 1: Linear Decomposer Networks (Rank-1 Autoencoders)}
In the first experiment, each subnetwork was parameterized by a \textit{rank-1 projection operator} of the form $u_i u_i^T$. This model is equivalent to a shallow autoencoder with a single latent scalar coefficient. The Decomposer Network was trained on the standardized AT\&T face dataset using the proposed iterative residual learning scheme and per-sample singular weights $\sigma_i$. 

\begin{figure}[h]
    \centering
    \includegraphics[width=0.95\linewidth]{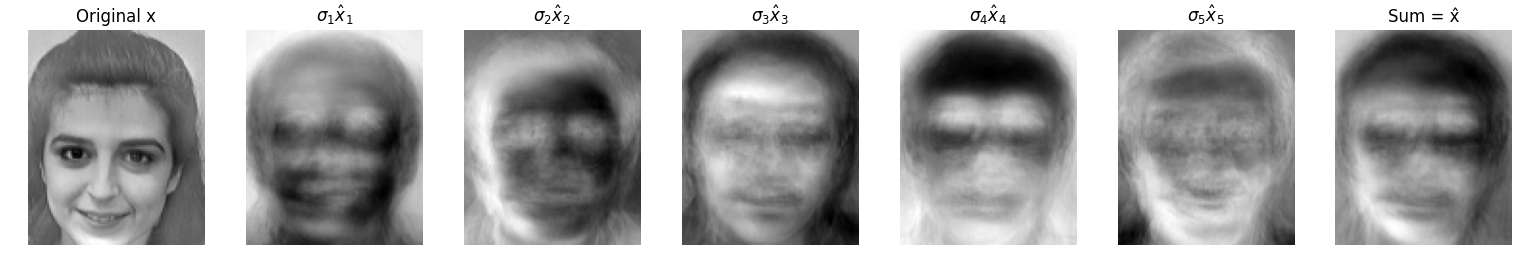}
    \caption{Experiment 1: Rank-1 linear subnetworks converge to PCA-like components on the AT\&T dataset. The learned components resemble the top singular vectors of the data matrix.}
    \label{fig:exp1}
\end{figure}

Despite being trained through gradient-based optimization, the learned projection directions $\{u_i\}$ converged to the principal directions of the dataset. This behavior is expected: under linearity and orthogonality constraints, the Decomposer Network minimizes the same objective as the Singular Value Decomposition (SVD) or Principal Component Analysis (PCA). As shown in Fig.~\ref{fig:exp1}, each component aligns with the dominant eigenvectors of the data covariance matrix, confirming that the architecture recovers PCA-like bases through unsupervised training. In three presented experiments, the first image is the original image and then five components are shown and the last image is combination of components.

\subsection{Experiment 2: Unconstrained CNN Autoencoders}
In the second experiment, the rank-1 restriction was removed and replaced with 3-layer convolutional autoencoders. Each subnetwork could now model nonlinear and spatially structured features. Without additional constraints, the subnetworks jointly learned overlapping but diverse reconstructions of the same input. 

\begin{figure}[h]
    \centering
    \includegraphics[width=0.95\linewidth]{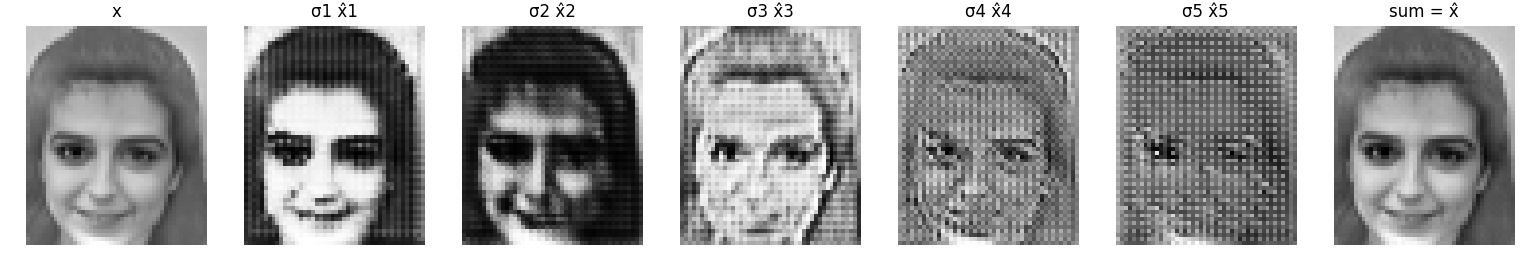}
    \caption{Experiment 2: CNN-based subnetworks without spatial constraints. Each component contributes differently to the reconstruction but all retain global image structure.}
    \label{fig:exp2}
\end{figure}

While the overall reconstruction $\hat{x} = \sum_i \sigma_i \hat{x}_i$ matched the input closely, the individual components $\hat{x}_i$ still exhibited global traces of the original face. This shows that, in the absence of explicit spatial or semantic disentanglement, the subnetworks collectively distribute the information but do not specialize in localized or interpretable features. The results, shown in Fig.~\ref{fig:exp2}, illustrate that the decomposition still captures multiple expressive modes of reconstruction even without orthogonality or region-based separation. 

\subsection{Experiment 3: Spatially Masked Decomposer Networks}
To encourage spatial specialization and semantic disentanglement, the third experiment introduced fixed \textit{Gaussian masks} before each autoencoder. The masks, defined over the image domain, were centered at random coordinates and designed such that each had a $0.5$-level contour covering approximately half the image area. These masks modulated the input residual for each subnetwork, guiding each one to focus on a specific spatial region while preserving overlap at the boundaries.

\begin{figure}[h]
    \centering
    \includegraphics[width=0.95\linewidth]{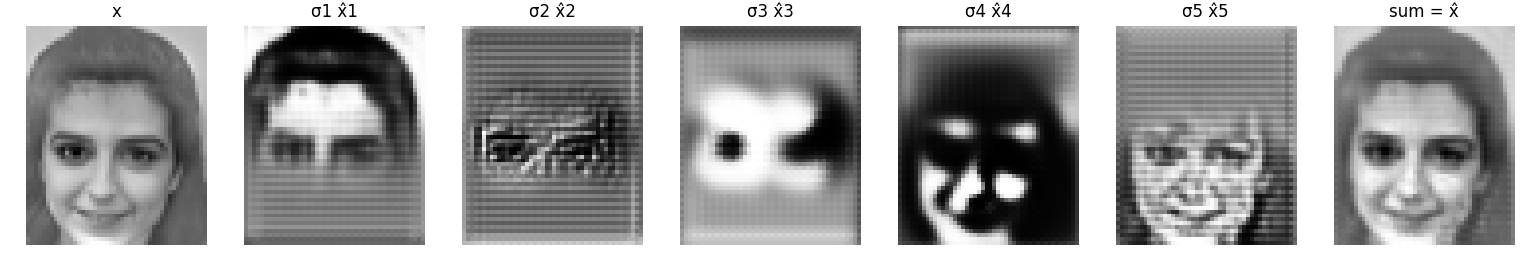}
    \vspace{-3mm}
    \caption{Experiment 3: Decomposer Networks with fixed Gaussian spatial masks. Each component captures a semantically meaningful subregion of the input image.}
    \label{fig:exp3}
\end{figure}

This modification resulted in more interpretable decompositions: individual subnetworks captured localized facial attributes such as eyes, mouth, or shading patterns, while the aggregated reconstruction $\hat{x}$ remained faithful to the original. As shown in Fig.~\ref{fig:exp3}, the decomposition became semantically meaningful that represents a coherent spatial or textural region within the face. This suggests that fixed masking can impose structured priors that lead to human-interpretable subcomponents without explicit supervision.

\subsection{Summary}
Across the three experiments, the proposed architecture demonstrated a progression from linear decomposition (recovering SVD/PCA) to nonlinear expressive components, and finally to semantically structured representations through spatial priors. This progression highlights the flexibility of Decomposer Networks as a unified framework bridging classic linear decomposition and modern deep feature factorization.

\section{DecompNet for Synthesis and Control}
\label{sec:synthesis}

Beyond analysis and decomposition, Decomposer Networks (\textbf{DecompNet}) naturally support \emph{controlled synthesis}.  
Each input $x$ is represented as a sum of learned semantic components, modulated by per-sample coefficients $\sigma_i$:
\[
\hat{x} = \sum_{i=1}^{N} \sigma_i \hat{x}_i, \qquad \hat{x}_i = f_i(r_i),
\]
where $f_i$ denotes the $i$th subnetwork and $r_i$ is its residual input.  
Since each component $\hat{x}_i$ corresponds to a coherent and interpretable substructure (spatial or conceptual), the coefficient $\sigma_i$ can be interpreted as a \textit{semantic control weight}.  
By modifying these weights after training, DecompNet can generate new samples that smoothly vary one semantic factor while keeping others fixed.

\subsection{Semantic Factor Manipulation}
In the linear case (Section~\ref{sec:relation_to_classic}), modifying $\sigma_i$ scales the contribution of the corresponding principal component, akin to classic PCA synthesis.  
In the nonlinear and masked configurations, however, each $\hat{x}_i$ represents a learned nonlinear generator for a specific semantic attribute.  
For instance, one subnetwork may implicitly encode global illumination, another may capture facial expression, and a third may represent background shading.  
Adjusting $\sigma_i$ for the ``illumination'' component allows us to \emph{brighten} or \emph{darken} the synthesized face without retraining the network or providing explicit attribute labels:
\[
x_{\text{synth}} = \sum_i \tilde{\sigma}_i \hat{x}_i, \quad 
\tilde{\sigma}_j \neq \sigma_j \text{ for illumination factor } j.
\]
This mechanism provides interpretable, low-dimensional control over the generated appearance while preserving image fidelity.

\subsection{Relation to Controllable and Disentangled Generation}
The concept of tuning $\sigma_i$ aligns closely with efforts in disentangled representation learning and controllable generative modeling.  
In particular, DecompNet shares conceptual similarities with:
\begin{itemize}
    \item \textbf{Disentangled VAEs} such as $\beta$-VAE~\cite{higgins2017beta} and FactorVAE~\cite{kim2018disentangling}, which separate latent factors but rely on global latent variables rather than structured residual pathways.
    \item \textbf{GAN-based control models} like StyleGAN~\cite{karras2019style}, where latent style vectors modulate specific layers to affect semantic attributes such as color or lighting.
    \item \textbf{Object-centric generative models} such as MONet~\cite{burgess2019monet} and IODINE~\cite{greff2019iodine}, which iteratively reconstruct image regions and allow slot-wise manipulation.  
          Unlike these models, DecompNet achieves component control without attention mechanisms or probabilistic inference; the control variable $\sigma_i$ is explicitly interpretable and directly tied to the reconstruction weights.
\end{itemize}

\subsection{Potential Applications}
This controllable synthesis property enables DecompNet to serve as a semantic editing framework.  
After training on natural images, one could modify $\sigma_i$ to:
\begin{itemize}
    \item Adjust lighting or shading by tuning an ``illumination'' component.
    \item Manipulate expression intensity while keeping identity constant.
    \item Combine components from different images to create hybrid compositions (e.g., swapping background vs.\ facial texture).
\end{itemize}
Such controllable synthesis bridges classical linear component editing (as in PCA morphing) and modern interpretable generative modeling, offering a deterministic, explainable alternative to latent-space manipulation in VAEs and GANs.

\subsection{Discussion}
The synthesis behavior of DecompNet underscores its dual role as both an \emph{analyzer} and a \emph{synthesizer}.  
Because each subnetwork learns to reconstruct a specific residual aspect of the input, the learned $\{\hat{x}_i\}$ act as basis generators, while the coefficients $\{\sigma_i\}$ form a semantic coordinate system.  
In contrast to typical deep generative models, these coordinates are not latent abstractions but physically interpretable scaling factors of identifiable visual components.  
This property opens avenues for zero-shot semantic editing and for data-driven control in creative or scientific image synthesis applications.

\section{Conclusion}
We introduced \textbf{Decomposer Networks}, a semantic autoencoder based on 
residual all-but-one factorization. This model brings together the interpretability 
of classical decomposition and the expressiveness of deep neural networks, 
opening a new path for semantic disentanglement in complex domains. Decomposer Networks extend the concept of singular vectors and singular values to deep components and their contribution. As DecompNet becomes shallower, the components merge to principal components defined by SVD.

\bibliographystyle{plainnat}
\bibliography{main}
\end{document}